# A Personalized Exercise Assistant using Reinforcement Learning (PEARL): Results from a four-arm Randomized-controlled Trial

Amy Armento Lee[1,2,*], Narayan Hegde[1,*], Nina Deliu[1,2,3,†], Emily Rosenzweig[1,4], Arun Suggala[1], Sriram Lakshminarasimhan[1], Qian He[1], John Hernandez[1], Martin Seneviratne[1], Rahul Singh[1], Pradnesh Kalkar[1], Karthikeyan Shanmugam[1], Aravindan Raghuveer[1], Abhimanyu Singh[1], My Nguyen[1], James Taylor[1], Jatin Alla[1], Sofia S. Villar[1,2,†], Hulya Emir-Farinas[1]

[1] Google, Mountain View, CA
[2] MRC Biostatistics Unit, Cambridge Institute of Public Health, School of Clinical Medicine, University of Cambridge, Cambridge, UK
[3] MEMOTEF Department, Sapienza University of Rome, Rome, Italy
[4] Ascension, St. Louis, MO

[†] Work done while at visiting researcher position at Google

*Correspondence: hegde@google.com and amyalee@google.com

# Abstract

Consistent physical inactivity among adults and adolescents poses a major global health challenge. Mobile health (mHealth) interventions, particularly Just-in-Time Adaptive Interventions (JITAIs), offer a promising avenue for scalable and personalized physical activity promotion. However, developing and evaluating such adaptive interventions at scale, while integrating robust behavioral science, presents methodological hurdles. The PEARL study was the first large-scale real world, four-arm randomized controlled trial designed to assess the effectiveness of a reinforcement learning (RL) algorithm informed by health behavior change theory, to personalize the content and timing of physical activity nudges via a Fitbit app.

We enrolled and randomized 13,463 Fitbit users to the study arms: control, random, fixed, and RL. Participants in the control arm received no nudges. The remaining arms all drew from the same bank of 155 nudges developed based on behavioral science principles, but according to different allocation mechanisms. The random arm participants received nudges whose content and time was selected at random. Participants in the fixed arm received nudges based on a fixed logic derived from their answers to survey questions designed to measure their barriers to physical activity. The RL group received nudges selected by an adaptive RL algorithm. Overall, 7,711 participants meeting the original inclusion criteria were included in primary analyses (mean age 42.1 years, 86.3% female, baseline daily steps 5,618.2). We observed an increase in physical activity (PA) for the RL group compared to each of the other groups, from baseline to 1 month and 2 months. The RL group had significantly increased average daily step count at 1 month compared to all other groups: the control group (+296 steps, p=0.0002), the random group (+218 steps, p=0.005), and the fixed group (+238 steps, p=0.002). At 2 months, the RL group sustained a significant increase as compared to the control group (+210 steps, p=0.0122). Furthermore, generalized estimating equation models revealed a sustained, significant increase in daily steps in the RL group compared to the control group (+208 steps, p=0.002). These findings demonstrate the potential of a scalable, behaviorally-informed RL approach to personalize digital health interventions for physical activity.

# Introduction

Currently, nearly one-third of adults and 80% of adolescents worldwide fail to meet WHO physical activity recommendations[1,2]. Inactivity rates vary significantly by region and demographics, with higher prevalence among women and in lower-middle-income countries. Alarmingly, global inactivity is projected to reach 35% by 2030[3]. The global rise in sedentary behavior has placed physical activity at the forefront of public health agendas, calling for scalable and pragmatic strategies. In particular, as emphasized by the Director of Health Promotion at WHO: "*We need to find innovative ways to motivate people to be more active, considering factors like age, environment, and cultural background.*"

There is growing experimental evidence that consumer wearable-based interventions using devices such as Fitbits are effective in increasing physical activity[4]. A comprehensive overview of recent mobile health and physical activity studies is presented in Table 1. By leveraging mobile devices like apps and wearables, mHealth can deliver timely, personalized interventions. Just-In-Time Adaptive Interventions (JITAIs) aim to provide scalable, accessible, context-specific support, adapting to individual needs, preferences, and fitness levels through continuous monitoring[5]. Through sophisticated algorithms, JITAIs can be dynamically adjusted based on user feedback, progress, and evolving behavior. Moreover, they can help identify and target specific population subgroups most vulnerable to inactivity and most likely to benefit from intervention programs.

To deliver on the full potential of JITAIs, it is paramount to balance their effectiveness with user burden. This entails effectively integrating into their design: (1) evidence-based behavioral theories, (2) experimental data from mHealth settings, and (3) adaptable, personalized intervention policies. This work reports and discusses the results of the first large-scale mHealth study that exemplifies this unique three-way implementation approach: the Personalized Exercise Assistant using Reinforcement Learning (PEARL) study. First, behavioral change theories are leveraged using the COM-B behavior change model[6] to categorize barriers to physical activity, and the Theoretical Domains Framework[7] to develop nudges that map on to these barriers. These models have been successfully applied across diverse health contexts, from medication adherence to physical activity interventions[8,9]. Second, the study design embeds micro-randomized trials (MRTs)[10], which enable researchers to examine the effectiveness of intervention components in real-time and detect any time-varying effects. Notably, this innovative approach may also help uncover patterns of engagement, and potential adverse effects such as habituation and retention. Finally, adaptable personalized interventions are integrated by utilizing machine learning algorithms, specifically, reinforcement learning (RL)[11]. In previous work, these algorithms have shown promising results across different health domains, including physical activity.

The PEARL study (https://osf.io/tw7up) was a randomized trial designed to assess the impact of different nudging strategies (including personalized), delivered as pop-up notifications via the Fitbit app, on daily step count. From February to June 2024, 13,463 adult Fitbit users (ages 22-60) in the U.S., who self-reported an average daily step count of less than 8,000 steps in the month prior to recruitment, were enrolled in the study. Among these, 7,711 were verified to have met the eligibility criteria and entered in the final analysis. Participants were randomized into one of four arms: a control (no nudge), and three intervention arms. The intervention arms all drew from the same predefined library of nudges, whose content was informed by behavioral science, and which were mapped to a set of theoretically derived barriers to physical activity. What differed between the arms was the method for selecting the message content and delivery timing for each participant on each day. One condition selected nudges and times of day at random. The second used a fixed logic derived from participants' answers to survey questions designed to measure their barriers to physical activity. The third used an adaptive RL strategy. The primary outcome measured was the change in average daily step count, comparing the final month of the study with the pre-enrollment baseline.

To our best knowledge, this study represents the first large-scale real world, four-arm randomized controlled trial designed to assess the effectiveness of RL informed by health behavior change theory, to personalize the content and timing of physical activity nudges via a Fitbit app. The literature review outlined in Table 1 highlights the unique contribution of the PEARL study to this space.

**Table 1. Literature review on ongoing and completed studies utilizing mHealth technologies and personalized intervention delivery (through MRT/RL) to promote physical activity.**

| Study | Study design and Objective | Proximal outcome | Study population, sample size, and duration | Methods | Interventions |
|---|---|---|---|---|---|
| **PEARL**<br>***Results*** | Four-arm RCT to increase physical activity:<br>- control group (no nudge)<br>- random group (daily random messages)<br>- fixed group (daily statically personalized messages)<br>- adaptive group (daily adaptively personalized messages) | Changes in daily step counts | N = 7,711 adult Fitbit users with <8,000 average daily steps<br><br>T = 60 days | - Fitbit tracker & App<br>- RL | - No message<br>- Five types of motivational messages: ability, perceived benefit, planning, prioritization, physical opportunity, social opportunity<br>- Timing of the message delivery daily (morning, evening) |
| Spruijt-Metz et al. (2022)<br>**HeartSteps 2**<br>***Protocol*** | Single-arm MRT to deliver JITAIs for managing physical activity and | Different according to the intervention type:<br>A. daily step count | N = 80 overweight, but otherwise healthy adults (planned) | - Contextual bandit RL<br>- Fitbit tracker & App | Three push intervention components:<br>A. motivational messages (standard daily |
| | sedentary behavior and build theoretical models in mHealth | B. 30-min step count following the decision time<br>C. time to sedentary-behavior-disruption | T = 90 days | | randomization)<br>B. walking suggestions (contextual bandit RL)<br>C. anti-sedentary suggestions (personalized algorithm multiple times per day) |
| Klasnja et al. (2019)<br>**HeartSteps 1**<br>***Results*** | Single-arm MRT to encourage regular walking via JITAIs | (Log)step count in the 30 min following randomization | N = 44 healthy sedentary adults (monetary compensation)<br><br>T = 42 days | - Jawbone Up Move tracker<br>- App | - No suggestion (0.4 randomization)<br>- Walking suggestion (0.3 randomization)<br>- Anti-sedentary suggestion (0.3 randomization) |
| Yom-Tov et al. (2017)<br>***Results*** | Two-arm study to increase the level of physical activity:<br>- control group (fixed reminders once-weekly)<br>- personalized group (daily feedback messages = weekly summaries) | Change in the activity (walking): minutes of activity on day t+1 divided by those on day t; rate of walking (steps per minute) | N = 27 type 2 diabetes patients<br><br>T = 26 weeks | - Phone accelerometer & App<br>- RL | - No message<br>- Three types of feedback messages: mastery, performance approach, and performance avoidance |
| Rabbi Mashfiqui et al. (2015)<br>**MyBehavior**<br>***Results*** | Single-arm N-of-1 study to promote physical activity and dietary health:<br>- phase 0 baseline: no suggestions;<br>- phase 1 control: random daily allocation;<br>- phase 2 experimental: RL | Changes in food calorie intake and calorie loss in exercise (log data) | N = 16 interested individuals identified through Cornell University's Wellness Center (monetary compensation)<br><br>T = 14 weeks | - App<br>- RL (Exp3) | - Daily food suggestions<br>- Daily activity suggestions |
| Zhou et al. (2018)<br>**mSTAR**<br>***Results*** | Two-armed RCT to increase physical activity:<br>- control group (fixed daily step goals of 10,000 steps per day)<br>- intervention | Relative change in daily steps from run-in to the 10-week follow-up | N = 64 adult staff employees of UCB (monetary compensation)<br><br>T = 10 weeks | - App<br>- RL | - Daily goal (steps target) |
| | group<br>(personalized daily step goals) | | | | |
| Zhou et al. (2018)<br>**mSTAR**<br>***Pilot results*** | Two-armed RCT to increase physical activity:<br>- control group (fixed daily step goals of 10,000 steps per day)<br>- intervention group (personalized daily step goals) | Number of steps | N = 13 UCB students (monetary compensation)<br><br>T = 10 weeks | - App<br>- RL (Inverse RL) | - Daily goal (steps target) |
| Aguilera et al. (2024)<br>**DIAMANTE**<br>***Results*** | Three-arm RCT to increase physical activity:<br>- control group (weekly mood monitoring message)<br>- random group (daily random messages)<br>- adaptive group (daily personalized messages) | Changes in daily step counts | N = 168 people with diabetes and depression symptoms (monetary compensation)<br><br>T = 24 weeks | - App<br>- RL | - Feedback messages (no message option)<br>- Motivational messages (no message option)<br>- Timing of the message delivery daily |
| Figueroa et al. (2022)<br>**DIAMANTE**<br>***Pilot study*** | Three-arm RCT to increase physical activity:<br>- control group (weekly mood monitoring message)<br>- random group (daily random messages)<br>- adaptive group (daily personalized messages) | Changes in daily step counts | N = 93 students at UCB (monetary compensation)<br><br>T = 6 weeks | - App<br>- RL | - Feedback messages (no message option)<br>- Motivational messages (no message option)<br>- Timing of the message delivery daily |
| Doherty et al. (2024)<br>**i80 BPM**<br>***Results*** | Randomized crossover trial (weekly basis) to increase exercising:<br>- control conditions (generic exercise sessions)<br>- RL conditions (personalized exercise sessions) | User satisfaction after each exercise session | N = 62 healthy adults recruited from Dublin and its environs via word of mouth and social media<br><br>T = 12 weeks | - App<br>- RL | - Exercise sessions |
| Lauffenburger et al. (2024)<br>**REINFORCE**<br>***Results*** | Two-arm RCT to improve adherence and personalize communication:<br>- control group (no messages)<br>- RL group (personalized messages) | Adherence to medication (fraction of pills taken) | N = 60 individuals with diabetes and glycated hemoglobin A1c [HbA1c] ≥ 7.5%<br><br>T = 6 months | - Text messages<br>- RL | - Daily text messages with 5 behavioural factors: (1) “Framing”: classified as neutral, positive, or negative, (2) observed feedback “History”: number of days in the prior week the patient was adherent, (3) “Social”, i.e., referring to loved ones, (4) “Content” reminder or informational, and (5) whether the text included a reflective question (“Reflective”) |
| Golbus et al. (2023)<br>**VALENTINE**<br>***Results*** | Two-arm RCT to deliver a JITAIs to promote physical activity in center-based cardiac rehabilitation patients:<br>- control group (no messages)<br>- intervention group (JITAIs) | Change in 6-min walk distance from baseline to 6-months | N = 220 low and moderate risk patients enrolled in center-based cardiac rehabilitation<br><br>T = 6 months | - Smartwatch<br>- Fitbit also included<br>- App | - Activity text messages - exercise text messages |

# Results

## Study population

Between February and June 2024, 15,000 Fitbit users were recruited through the Fitbit app's Steps dashboard. Following eligibility screening, 13,463 users were enrolled and randomly assigned to one of four groups for a 60-day follow-up: a control group (n=3,304, 24.5%), a random nudge group (n=3,337, 24.8%), a fixed nudge group receiving nudges selected based on an entrance survey (n=3,419, 25.4%), and a personalized reinforcement learning group (n=3,403, 25.3%) (Figure 1).

Of the 13,463 enrolled participants, 7,711 (68.0%) participants met the criteria required for analysis of step-count, including meeting study eligibility criteria of less than 8,000 steps per day at baseline and availability of step-count data through the 60 day study period (Figure 1).

Baseline demographic characteristics are reported in Table 2, with no notable differences across the arms. Eligible participants had a mean age of 42.1 (SD = 9.0) years, with 6,657 women (86.3%). Mean body weight was 90.8 (SD = 25.5) kg, and baseline average daily steps was 5,618.2 (SD = 1,502.5).

**Figure 1. Consort diagram**

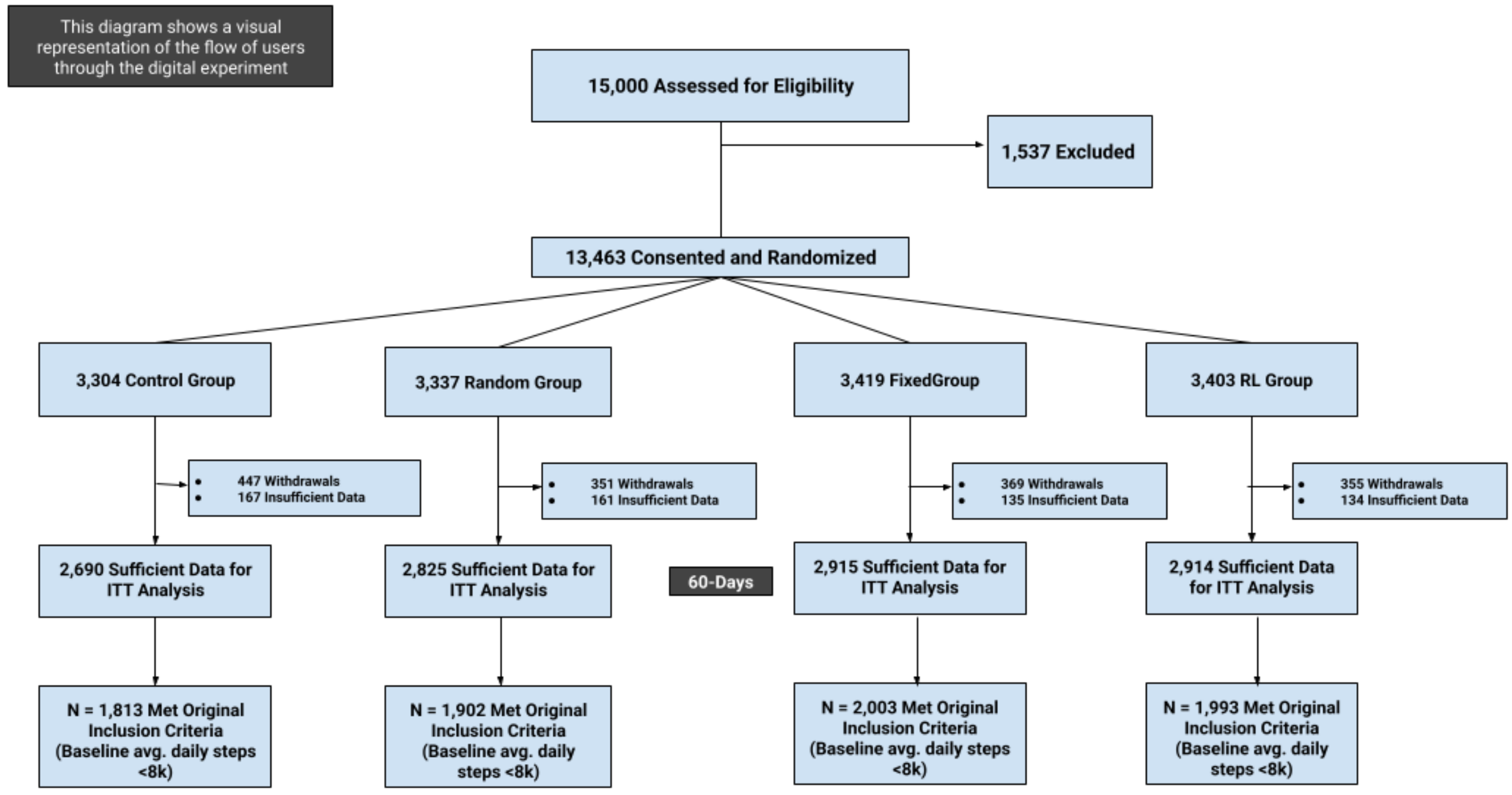

**Table 2. Baseline characteristics of study population (N = 7,711), reported as mean (SD) or N (%)**

| | Control<br>N = 1,813 | Random<br>N = 1,902 | Fixed<br>N = 2,003 | RL<br>N = 1,993 |
|---|---|---|---|---|
| Age (years) | 42.0 (9.1) | 42.1 (9.0) | 42.0 (8.8) | 42.2 (9.0) |
| Female | 1,578 (87.0%) | 1,641 (86.3%) | 1,705 (85.1%) | 1,733 (87.0%) |
| Weight (kg) | 91.1 (29.5) | 90.8 (23.6) | 90.9 (24.9) | 90.4 (23.8) |
| Baseline average daily steps | 7,248 (3,064) | 7,160 (3,004) | 7,170 (3,089) | 7,171 (3,073) |
| Phase / Month of Enrollment | | | | |
| Q1 2024 (Feb/March) | 691 (38.1%) | 766 (40.3%) | 833 (41.6%) | 840 (42.2%) |
| Q2 2024 (April, May, June) | 1,122 (61.9%) | 1,136 (59.7%) | 1,170 (58.4%) | 1,153 (57.9%) |
| Location (Self-report in Onboarding Survey) | | | | |
| Urban | 261 (14.4%) | 295 (15.5%) | 311 (15.5%) | 301 (15.1%) |
| Suburban | 1,153 (63.6%) | 1,216 (63.9%) | 1,267 (63.3%) | 1,240 (62.2%) |
| Rural | 376 (20.7%) | 375 (19.7%) | 408 (20.4%) | 435 (21.8%) |

## Study execution

Among 13,463 participants enrolled, 1,522 (11.3%) withdrew from the study (Figure 1), with a higher withdrawal rate in the control group (N = 447, 13.5%). Although participants were blinded to treatment allocation, the consent process indicated that group assignment would influence the type of messages received. The absence of any message delivery to the control group throughout the study period is very likely to have unblinded participants in this arm, potentially contributing to the observed numerically higher rate of withdrawal compared to the intervention group.

Participants were only included in the analysis if they met the following minimum requirements for data collection: for baseline and each study month, the participant must have at least 7 days with at least 500 steps per day. We use this threshold as a proxy for wear-time of the Fitbit wearable device, and expect that if a participant wore the device for a significant portion of the day there would be at least 500 steps. No imputation was applied for missing data for participants that were included in the analysis.

## Study outcomes

We analyzed the average daily step count in month 1 and month 2 compared to the month before enrollment (baseline month). Overall, 7,711 participants were included in the final analysis (Figure 1).

At the pre-study baseline month, the average daily steps amounted to 5,618.2 with no important differences across study arms (Table 3). At month 1, average daily step count increased to 6,050 (SD 1,922.0) in the control group, 6,061.2 (SD 1,897.5) in the random group, 6,054.8 (SD 1,875.9) in the fixed group, and 6,282.2 (SD 1,956.0) in the RL group. At month 2, average daily step count increased to 5,948.6 (SD 2,076.9) in the control group, 5,947.6 (SD 2,068.8) in the random group, 5,969.4 (SD 2,016.1) in the fixed and 6,096.3 (SD 2,117.4) in the RL group.

**Table 3. Step count across study arms, at each phase, reported as mean (SD).**

| Study Arm | Pre-Study Baseline Month | Month 1 | Month 2 |
|---|---|---|---|
| Control Group | 5,659.8 (SD = 1,491.6) | 6,050.0 (SD = 1,922.0) | 5,948.6 (SD = 2,076.9) |
| Random Group | 5,594.2 (SD = 1,503.8) | 6,061.2 (SD = 1,897.5) | 5,947.6 (SD = 2,068.8) |
| Fixed Group | 5,615.2 (SD = 1,515.1) | 6,054.8 (SD = 1,875.9) | 5,969.4 (SD = 2,016.1) |
| RL Group | 5,606.3 (SD = 1,498.7) | 6,282.2 (SD = 1,956.0) | 6,096.3 (SD = 2,117.4) |
| Grand Total | **5,618.2 (SD = 1,502.5)** | **6,114.0 (SD = 1,915.2)** | **5,991.9 (SD = 2,070.4)** |

Difference in differences regression[12] analysis was conducted to compare change in step count across treatment groups, results are presented in Table 4 (see Methods section 'Statistical Analysis' for more detail). Notably, the RL group demonstrated a statistically significant higher increase in steps at month 1 compared to all other groups. At month 2, the RL group had a higher increase in step count compared to all groups, this difference was only statistically significant for the comparison with the control group. The primary endpoint was to assess the difference in average daily step count change at month 2 for four specific pairwise comparisons, and these are indicated by bolding of the results in Table 4.

**Table 4. Difference in differences regression model results for average daily step count at month 1 and month 2 (N = 7,711)**

| | | Month 1 vs. Pre-study Baseline Month | Month 2 vs. Pre-study Baseline Month |
|---|---|---|---|
| Comparisons to Control Group | Random vs. Control | B = 77.6<br>SE = 79.7<br>P = 0.330 | **B = 63.1**<br>**SE = 84.0**<br>**P = 0.452**<br>**Adjusted P = 0.602** |
| | Fixed vs. Control | B = 57.5<br>SE = 78.8 | B = 75.1<br>SE = 82.7 |
| | | P = 0.465 | P = 0.364 |
| | RL vs. Control | B = 295.9<br>SE = 79.6<br>P = 0.0002*** | B = 210.1<br>SE = 83.8<br>P = 0.0122* |
| Comparisons to Random Group | Fixed vs. Random | B = -20.1<br>SE = 77.7<br>P = 0.796 | **B = 12.0**<br>**SE = 81.7**<br>**P = 0.883**<br>**Adjusted P = 0.883** |
| | RL vs. Random | B = 218.3<br>SE = 78.5<br>P = 0.005** | **B = 147.0**<br>**SE = 82.8**<br>**P = 0.076**<br>**Adjusted P = 0.195** |
| Comparisons to Fixed Group | RL vs. Fixed | B = 238.4<br>SE = 77.6<br>P = 0.002** | **B = 135.0**<br>**SE = 81.5**<br>**P = 0.098**<br>**Adjusted P = 0.195** |

*Bold indicates the primary analyses reported above. Benjamini-Hochberg adjustment used to control for Type 1 error among primary outcomes.*

**Figure 2. Trajectory of average daily step count per group (N = 7,711)**

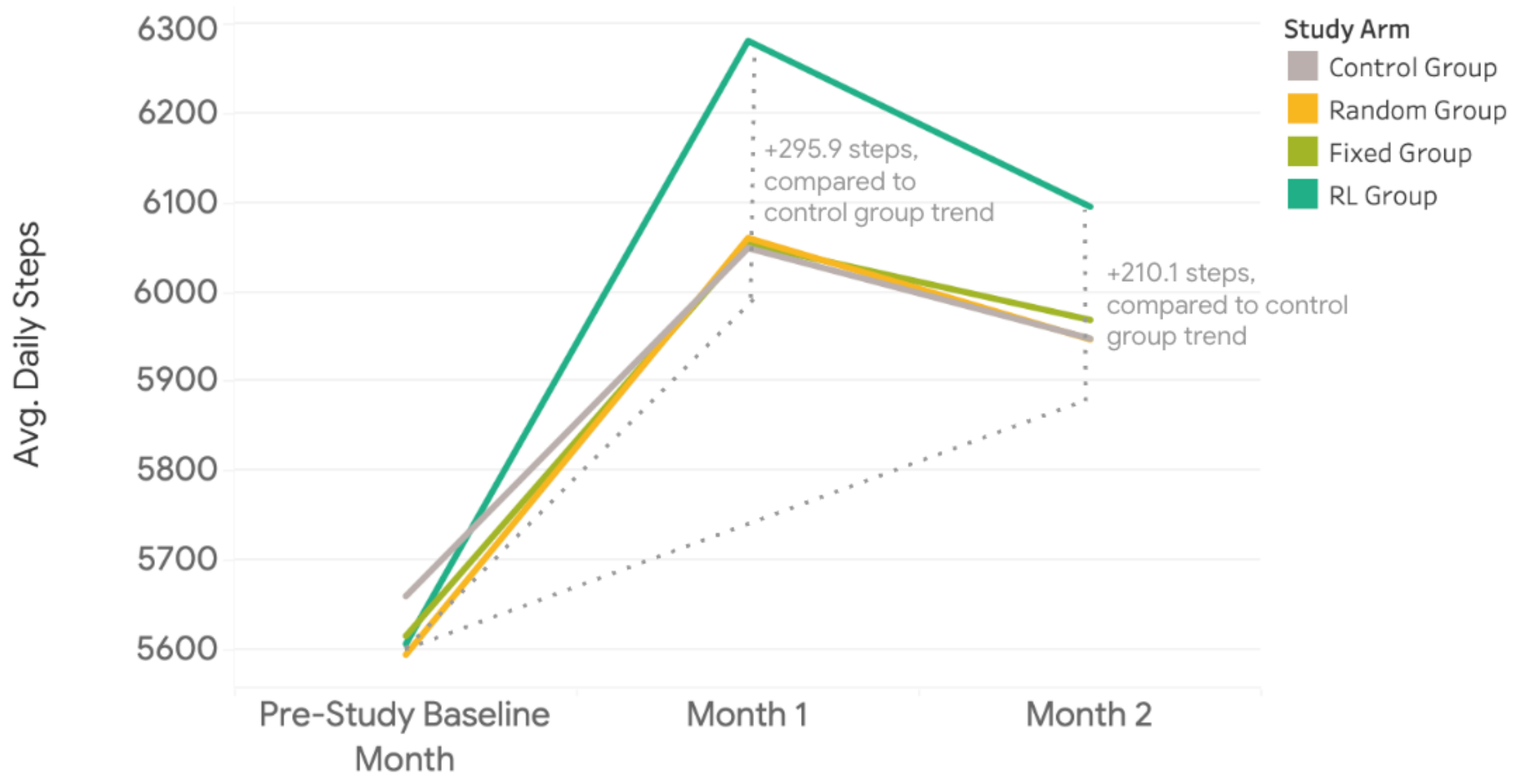

## Modelling effects over time

A generalized estimating equation model (GEE; exchangeable correlation structure)[13,14] was used to estimate the average effect of treatment group on the daily step count over time; the results are available in Table 6.

Compared to the control group, the fixed group demonstrated decreased average daily step count by -18.7 (SE = 67.9, p = 0.783); this effect increased linearly over time, by a step change of +1.29 (SE = 1.4, p = 0.373) compared to the control group. The RL group showed an increase in average daily step count of +208.0 (SE = 68.6, p = 0.002), compared to the control group; this effect diminished linearly over time, by a step change of -0.6 (SE = 1.4, p = 0.696) compared to the control group. The time effect was negative, with a decrease of -4.5 steps per day, this is consistent with other literature[15].

**Table 6. Results of the GEE model (N = 7,711)**

| Covariate | Estimate | 95% CI | p-value |
|---|---|---|---|
| (Intercept) | 6125.2 | (6028.1, 6222.4) | <0.001 |
| Treatment Arm: Random | -14.1 | (-148.1, 119.8) | 0.836 |
| Treatment Arm: Fixed | -18.7 | (-151.6, 114.3) | 0.783 |
| Treatment Arm: RL | 208.0 | (73.6, 342.4) | 0.002 |
| Day in study | -4.5 | (-6.6, -2.5) | <0.001 |
| Treatment Arm: Random x study day | 0.5 | (-2.3, 3.4) | 0.707 |
| Treatment Arm: Fixed x study day | 1.3 | (-1.5, 4.1) | 0.373 |
| Treatment Arm: RL x study day | -0.6 | (-3.4, 2.2) | 0.696 |

**Figure 3. Modelling longitudinal effect by study day (GEE) (N = 7,711)**

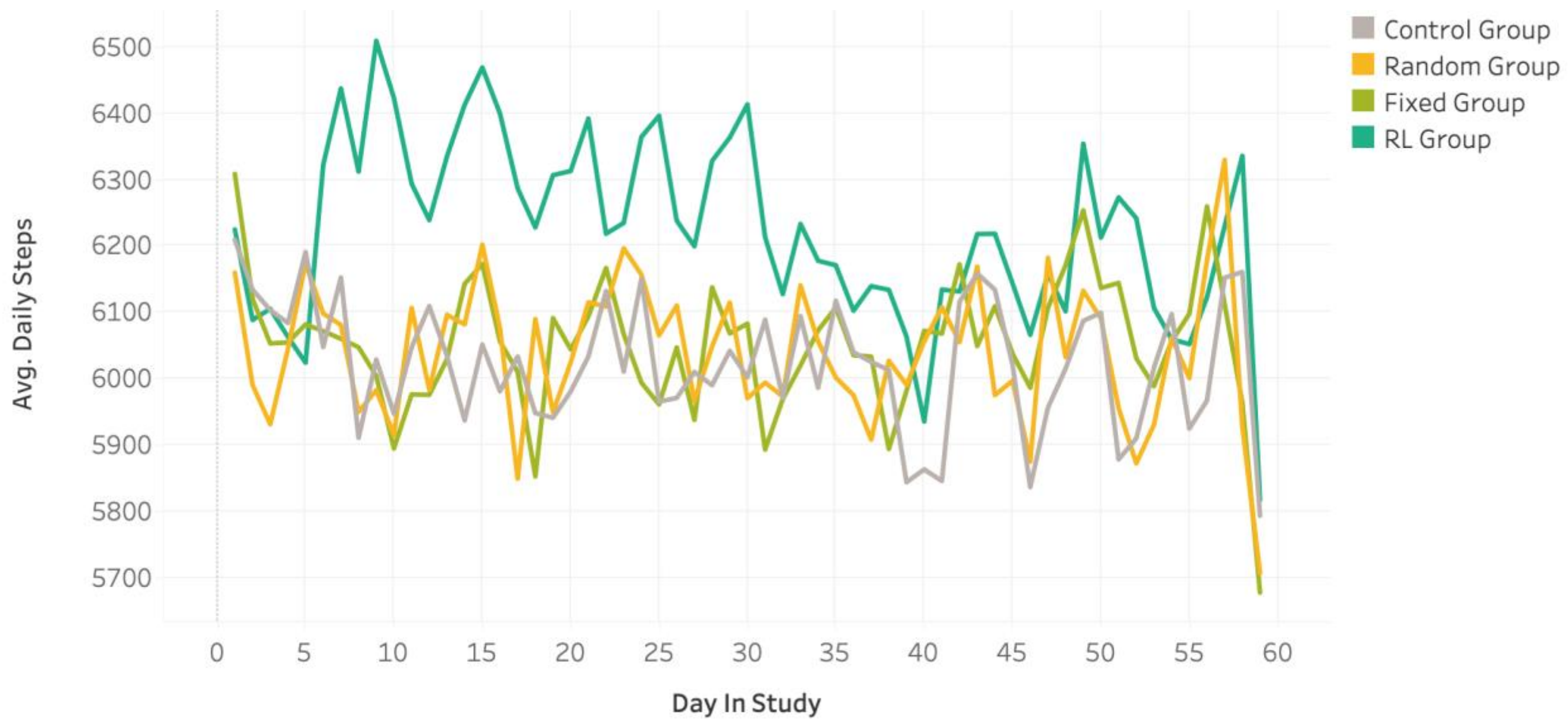

## COM-B Survey

Participants were asked to complete the COM-B survey during study onboarding and in the exit survey during the final week of the study (see the Methods section ‘COM-B Survey’ for more detail). This survey was designed to measure participants’ baseline levels of Capability, Opportunity, and Motivation, each of which contained two sub-constructs. Capability was divided into Ability and Planning, Opportunity was divided into Physical and Social, and Motivation was divided into Perceived Benefit and Prioritization. and EAlmost all participants completed the onboarding COM-B survey (99.8%), but only a fraction completed the exit COM-B survey (26.7%).  On average, the study cohort demonstrated the highest scores (reflecting the greatest levels of) Perceived Benefit and Ability, and the lowest scores for the themes Planning and Social Opportunity.

All of the nudges were classified as relating to / acting on one of these six themes. In the fixed condition, participants who scored lower on a given theme received more nudges associated with addressing that type of barrier to activity.

**Figure 4. COM-B survey questionnaire at study onboarding, mean & SD (N = 7,695)**

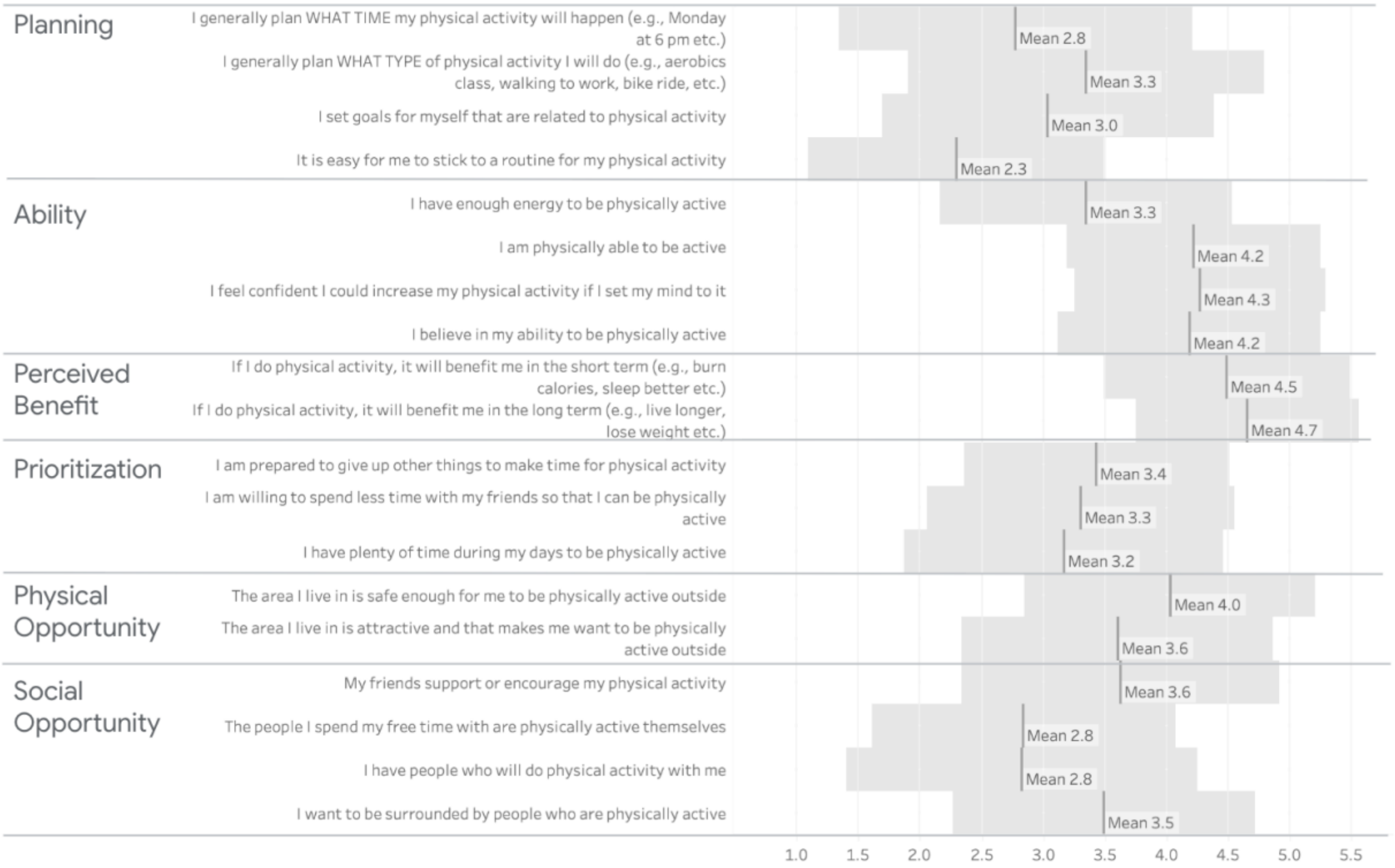

**Figure 5. Change in COM-B survey questionnaire, from baseline to study exit, mean & SD (N = 2,059)**

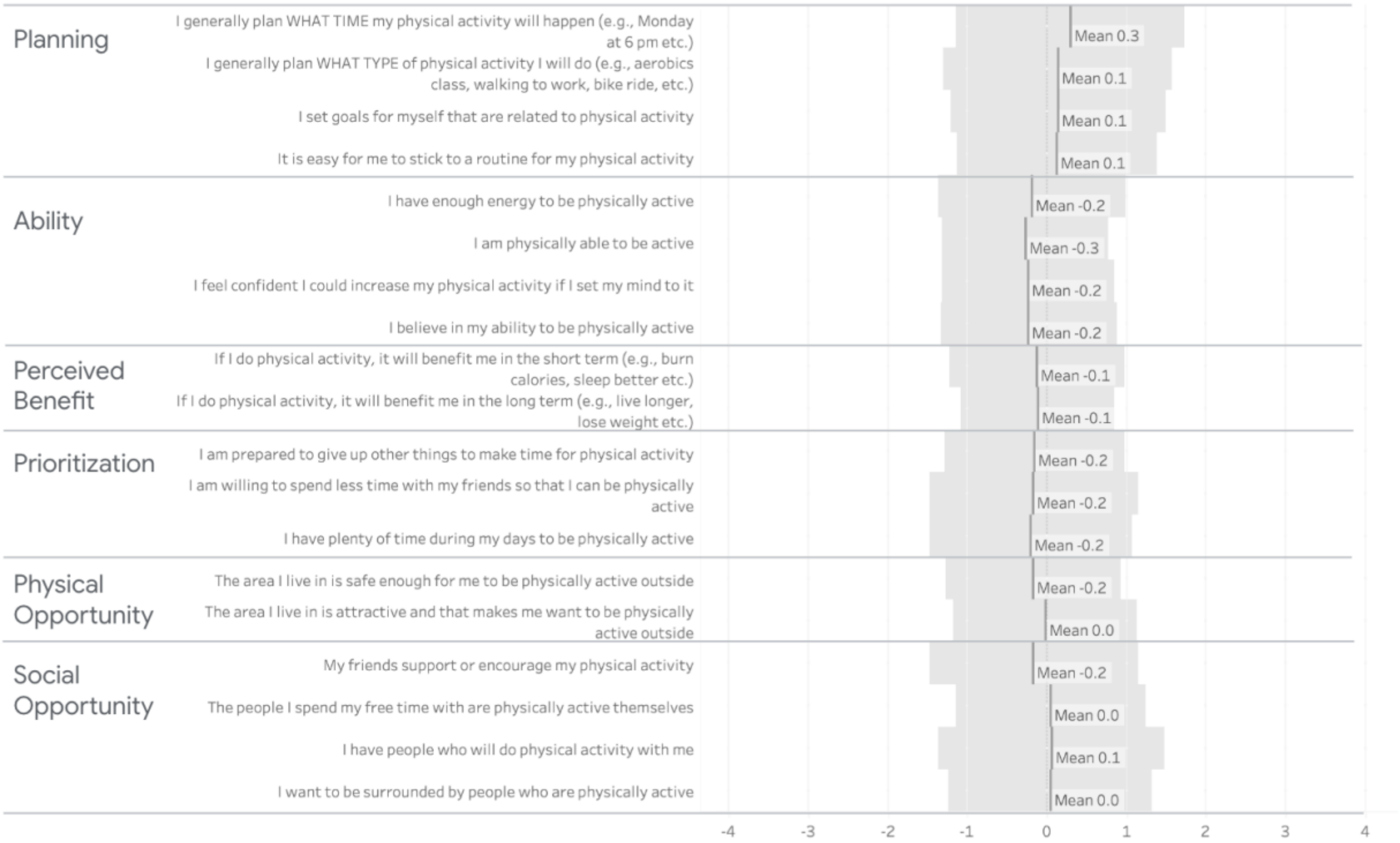

## Real-time User Feedback for Individual Nudges

User feedback was collected at the point of nudge delivery through optional thumbs-up/thumbs-down ratings and open-text comments. Thumbs-up and thumbs-down feedback was aggregated by the nudge COM-B theme. Nudges categorized under the Ability and Perceived Benefit themes received higher favorability rates, while those in the Physical Opportunity theme showed markedly lower favorability (Figure 8). Nudge feedback was included in the RL model (see Methods section 'Reinforcement Learning Policy'), but was not found to be among the most important features (Figure 12).

**Figure 8. Individual nudge feedback, aggregated by content type (COM-B theme)**

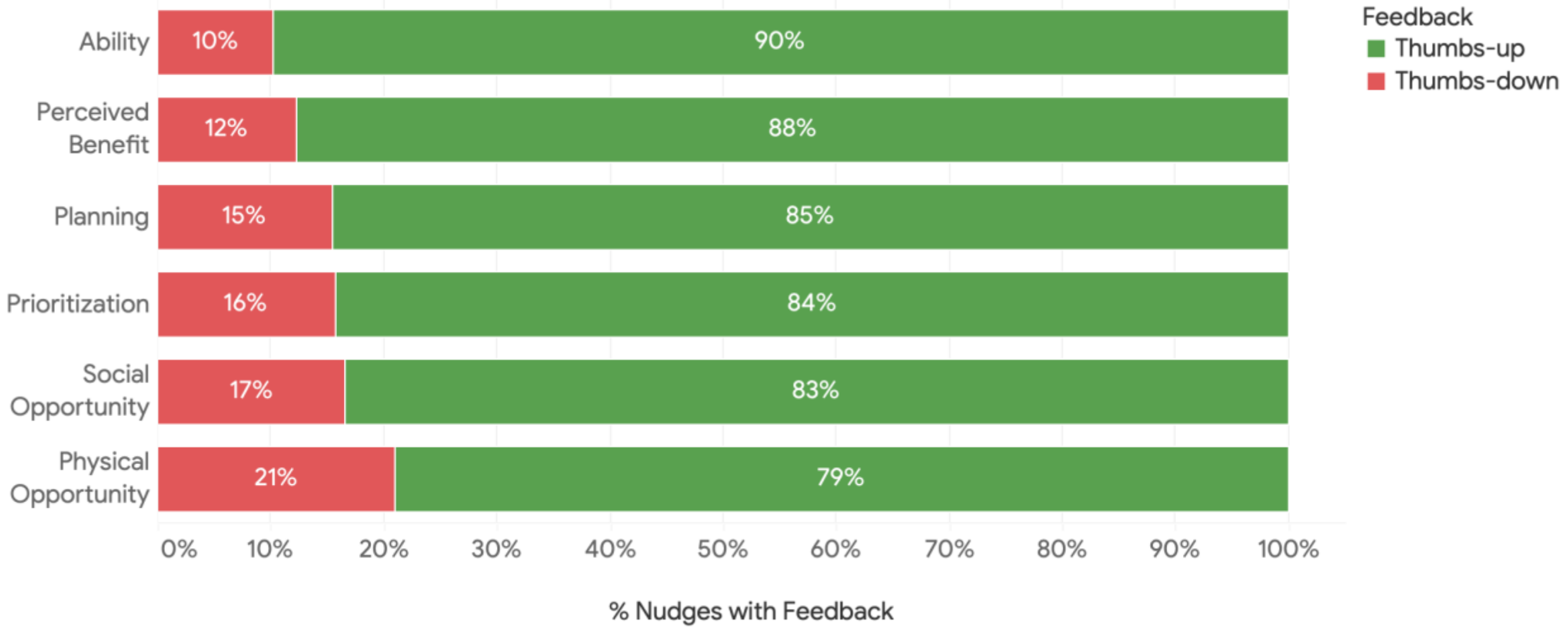

## User Feedback in Exit Survey

In the final week of the study, participants were invited to complete a study exit survey. In total, 2,151 (27.9%) participants completed the survey, with lower response rate in the no nudge group (17.9%). When asked "Would you recommend nudges to your friends/family starting on a fitness journey?", we observed a higher proportion of positive responses in the RL group (77.6%) compared to all other groups (48.9% for the control group, 74.0% for the random group, and 70.6% for the fixed group) (Figure 6).

**Figure 6. Exit survey questionnaire: would recommend nudges**

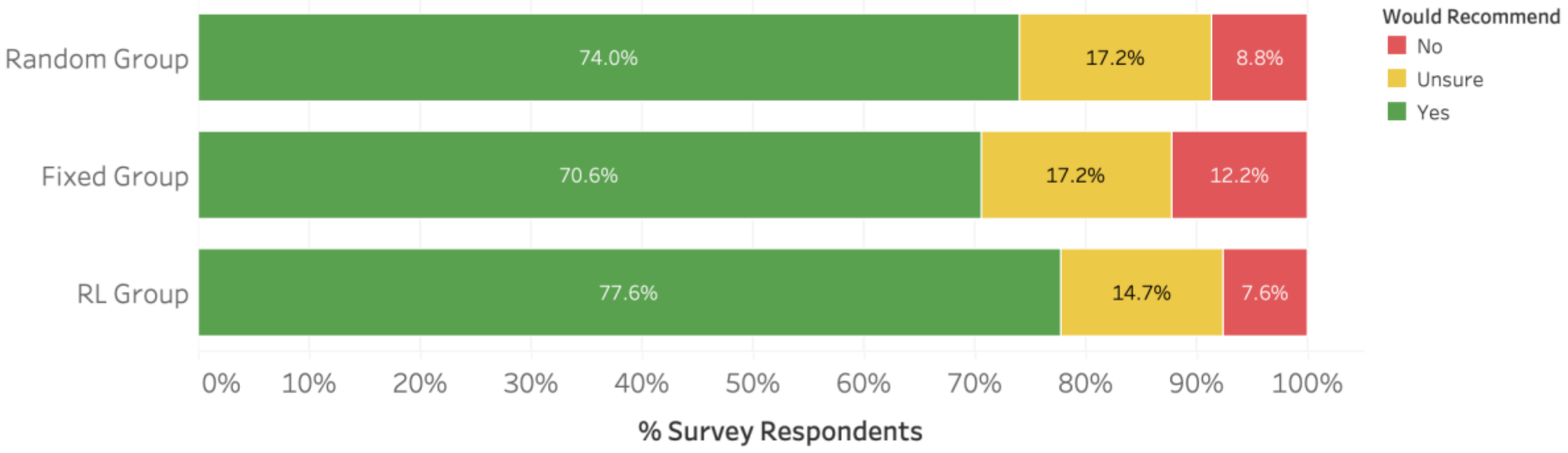

The exit survey also asked participants: “How would you rate the level of customization and relevance of nudges?”. Again, the RL group demonstrated higher rates of favorable responses (37%) as compared to all other groups (10.8% for the control group, 26.7% for the random group, 24.5% for the fixed group) (Figure 7). For both exit survey questions, the control group responded with non-zero rates of positive sentiment, despite receiving no nudges during the study period.

**Figure 7. Exit survey questionnaire: nudge customization and relevance**

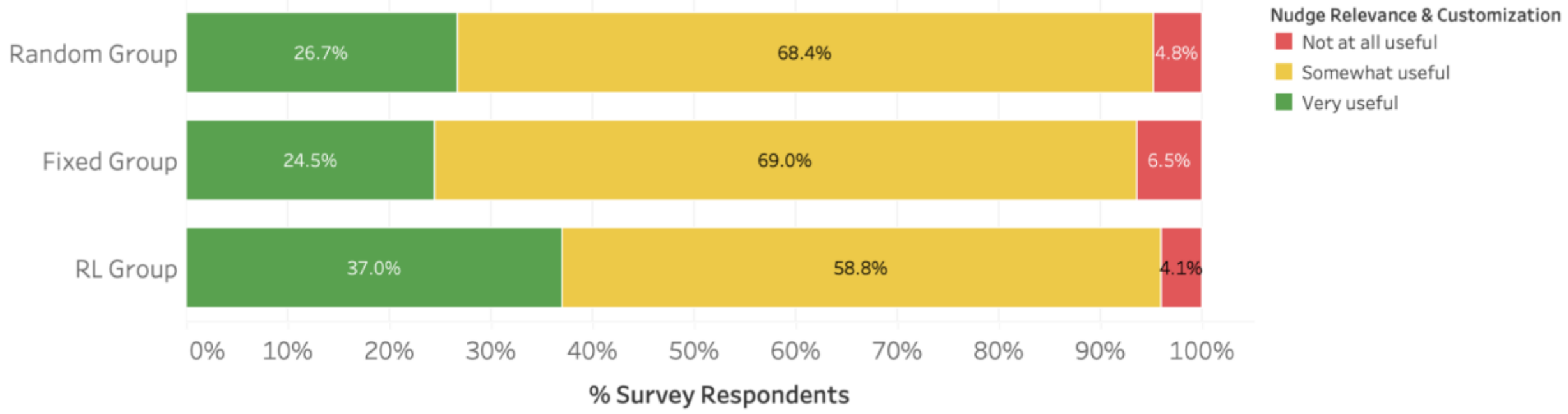

## Exploratory analyses: reinforcement learning group and personalization

The RL policy for nudge timing and content selection demonstrated distinct patterns compared to the random and fixed arms at the population level. Compared to the random group, the RL group delivered 10% more Ability themed nudges (Figure 9), and 7% more evening nudges (Figure 10); note that Ability themed nudges also received the highest favorability rates in the nudge-level feedback (Figure 8). The fixed group delivered the highest rates of Planning and Physical Opportunity themed nudges (Figure 9), though Physical Opportunity theme received the highest rate of negative nudge-level feedback (Figure 8).

**Figure 9. Distribution of nudge content across treatment groups**

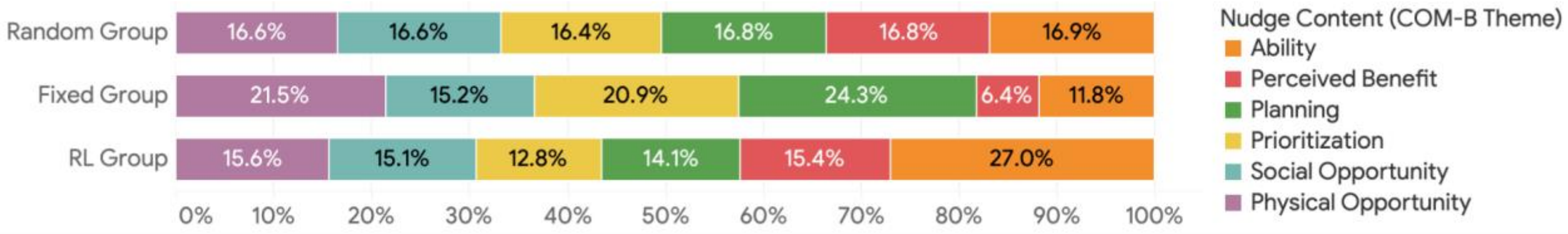

**Figure 10. Distribution of nudge timing across treatment groups**

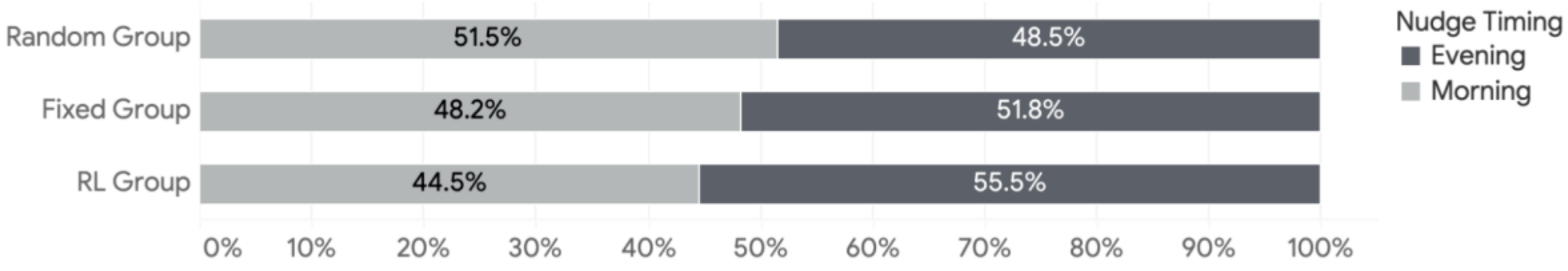

Analysis of nudge distribution over time indicates that the RL policy initially prioritized Perceived Benefit and Planning themes, shifting to the Ability theme in later stages, demonstrating temporal adaptation of the policy (Figure 11).

**Figure 11. Distribution of nudge content and timing selection across treatment groups, and over time**

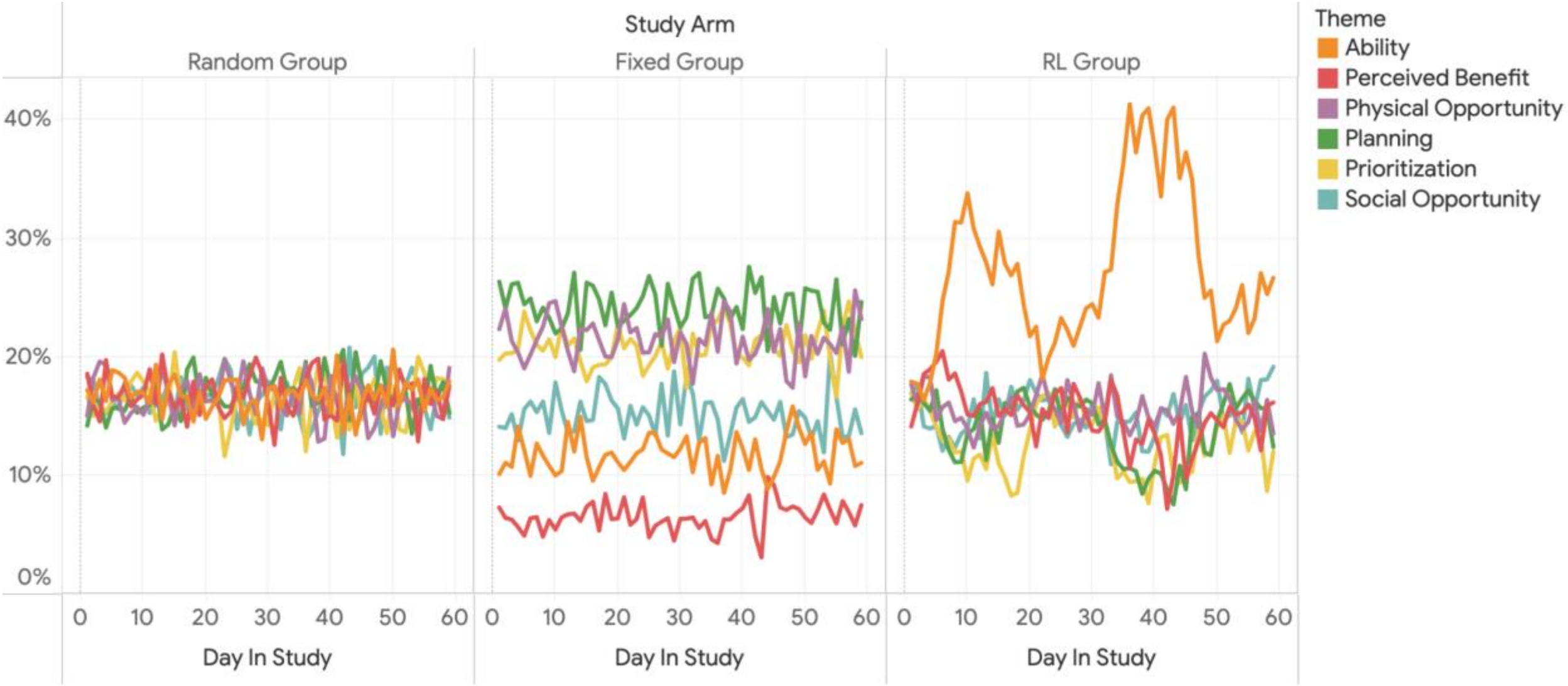

# Description of the reinforcement learning algorithm learning process

The most influential features and interactions from the reinforcement learning algorithm across the entire study duration are shown in Figure 12. Fixed characteristics that carried the most weight within the model were pre-study baseline step patterns, weight, and age. The behavioral pattern with the largest weight was the previous day step count. We refer to the Methods section (Reinforcement Learning Policy) for complete details on the features included in the model.

**Figure 12. Most influential features in the RL model (model weights)**

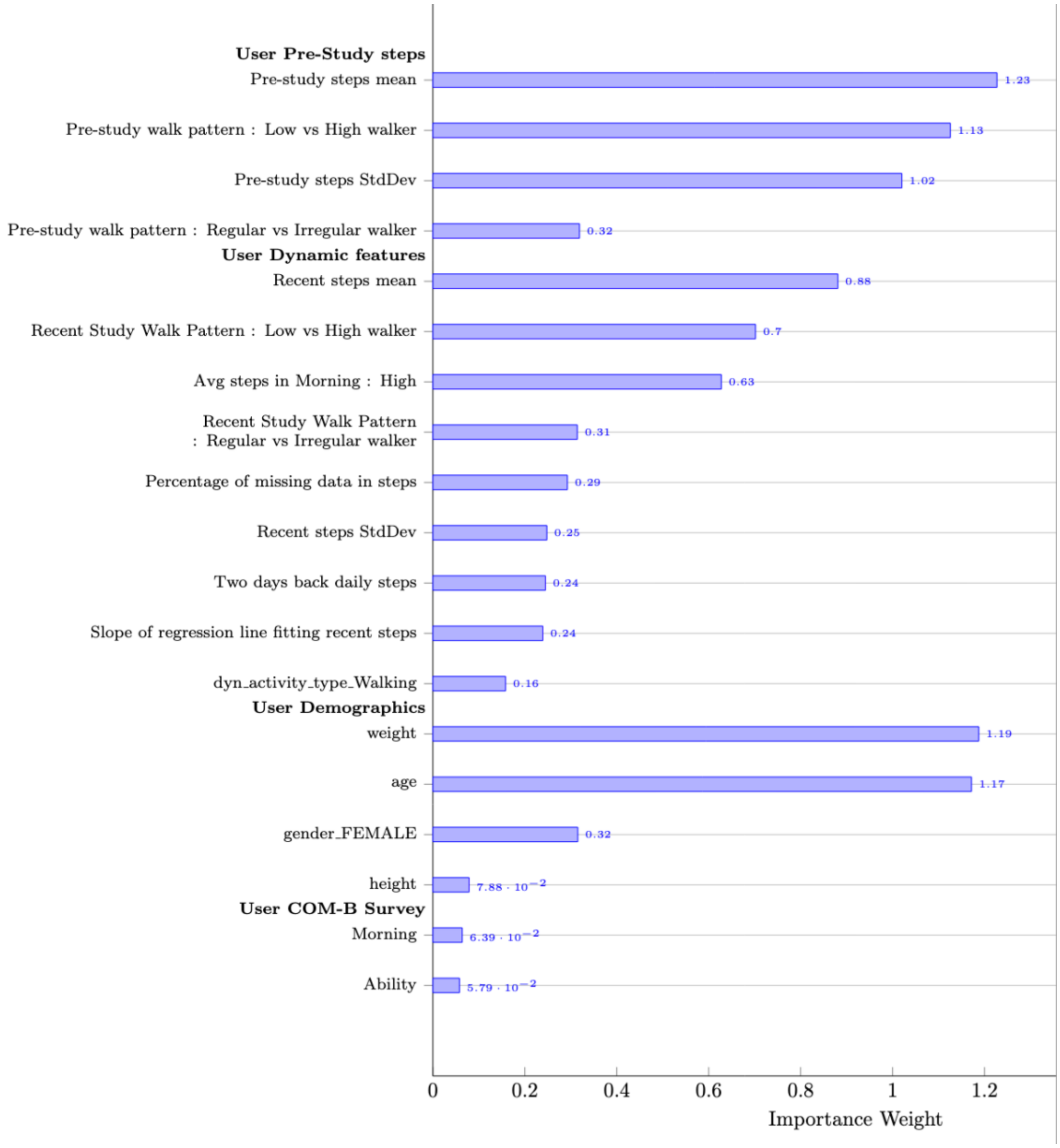

# Discussion

In this large-scale randomized controlled trial, we investigated the effectiveness of a novel reinforcement learning algorithm guided by the COM-B behavioral change model, in promoting physical activity through personalized nudges delivered via the Fitbit app. We observed an increase in average daily step count at month 1 and month 2, comparing the RL group to control, random, and fixed groups. This, along with the longitudinal modeling, reveal promising insights into the potential of adaptive, behaviorally-informed interventions.

## Contextualizing the Observed Step Changes

The significant increase in average daily steps observed in the RL group at one month (mean change +295.9 steps vs. control, p=0.0002; +238.4 steps vs. fixed, p=0.002) and the sustained, significant effect over time in the generalized estimating equation (GEE) model (+208.0 steps vs. control, p=0.002) are noteworthy. While the clinical meaningfulness of step count increases can vary, a consistent increase of 200-300 steps per day, if sustained over time, can contribute to improved health outcomes, particularly in sedentary populations[16]. For instance, studies have shown that even modest increases in daily steps are associated with reduced risks of cardiovascular disease and all-cause mortality[17]. This suggests that while the effect size might appear modest, its cumulative impact on a large population could be substantial, aligning with public health goals of reducing global inactivity.

## Effectiveness of the Fixed Arm

Our findings reveal that participants in the fixed arm, who received nudges selected based on a fixed logic relating to their baseline levels of Capability, Opportunity, and Motivation, did not demonstrate a significant advantage over the group whose nudges were selected at random. This outcome highlights a critical learning point for designing just-in-time adaptive interventions (JITAIs): a static, rule-based approach, even when informed by robust behavioral theory, may be insufficient in dynamic, real-world mHealth environments.

The PEARL study exemplified this challenge. A notable observation was the discordance between the high rate of Physical Opportunity nudges delivered by the fixed group (Figure 9) and the predominantly negative user feedback received for this theme (Figure 7). Open-text feedback suggests that Physical Opportunity nudges often relied on inaccurate assumptions about participants' real-world environments (e.g., availability of sidewalks or stairs). This underscores a key challenge in applying the baseline COM-B survey to dynamic, in-the-moment interventions. The COM-B framework, as typically applied, may not fully capture the day-to-day or context-specific fluctuations in behavioral barriers (e.g., work vs. home, weekday vs. weekend).

Human behavior is inherently contextual and non-stationary. A fixed intervention policy, determined at baseline, struggles to adapt to evolving user needs, preferences, and external circumstances (e.g., changes in daily routines, weather, motivation levels). This finding points to a gap in the current literature regarding the optimal translation of comprehensive behavioral frameworks like COM-B into continuously adaptive digital nudges. While COM-B effectively diagnoses behavioral determinants, the challenge lies in dynamically selecting and delivering the right intervention component (e.g., targeting Ability versus Opportunity) at the opportune moment for a given individual in real-time. Our results suggest that a dynamic algorithm capable of continuous learning and adaptation is crucial for leveraging behavioral science effectively in ongoing digital interventions. Future work should focus on adapting the COM-B framework to account for such temporal and contextual variability, potentially moving beyond general delivery policies to more deeply personalize nudge content by leveraging richer participant information.

## Strengths of the PEARL Reinforcement Learning Approach

In contrast to the fixed arm, the RL arm demonstrated a more sustained positive impact on physical activity. This strength stems from RL's inherent ability to learn and adapt over time, balancing exploration (trying different nudges) and exploitation (delivering nudges that have historically worked well for similar contexts/users). The detailed exploratory analyses within the RL arm further illustrate this adaptive capacity.

Furthermore, as highlighted in Table 1, most existing studies on personalized digital health interventions for physical activity are small-scale, with sample sizes typically ranging from 13 to 220 individuals and a median below 100. These studies frequently call for larger, longer-term investigations to fully evaluate the potential of digital health to significantly improve health-related behaviors and outcomes. PEARL directly addresses this gap by providing insights from a population of more than 7,000 participants followed over two months, making it one of the largest RL-based studies in physical activity and beyond.

Beyond its unprecedented scale, the PEARL study offers several key advantages. It features a unique four-arm randomized controlled trial design, which includes two embedded micro-randomized trials (MRTs), allowing for both overall treatment efficacy assessment and real-time evaluation of intervention components. Unlike some previous studies that relied solely on mobile apps, PEARL leveraged the continuous data collection from Fitbit wrist-worn trackers, providing a more robust measure of physical activity. This robust study design, combining traditional RCT elements with traditional behavior science interventions (random and fixed arms) and an augmented adaptive intervention, enables a deeper understanding of the effects. The ability of the RL algorithm to learn population-level preferences (e.g., evening nudges being more effective than morning nudges across themes) and then personalize further based on individual trajectories is a key differentiator. This adaptive nature, coupled with the large-scale deployment, positions the PEARL study as one of the largest and most comprehensive RL-based intervention studies in digital health to date, addressing critical challenges in real-world human subject research and large-scale data-driven interventions.

## Methodological Considerations and Generalizability

The study's large sample size and real-world recruitment through the Fitbit app are significant strengths. Compared to previous studies that primarily involved university students and staff[15,18], who may not be representative of the broader population, PEARL's sample may have higher generalizability. Furthermore, unlike most existing studies, PEARL did not offer monetary compensation to participants, minimizing potential selection bias. However, several factors warrant consideration regarding generalizability. The participant population was predominantly female (86.3%) and self-selected as Fitbit users, potentially limiting the direct applicability of findings to broader, more diverse populations, including men or individuals without wearable devices. Future research should aim to recruit more diverse cohorts.

The rolling enrollment between February and June 2024 introduces potential seasonality effects, as physical activity patterns can vary with weather and daylight hours. While baseline characteristics were balanced across groups, the impact of enrollment month as a predictor in subgroup analysis was not explicitly significant for the primary outcome, suggesting that randomization effectively distributed this potential confounder. Nevertheless, future studies could consider stratifying recruitment by season or incorporating seasonal variables into the RL algorithm.

The potential for unblinding in the control group, due to the complete absence of messages, is a limitation. While participants were blinded to their specific treatment allocation, the lack of any interaction in the control arm might have led to higher withdrawal rates or altered engagement compared to the intervention arms. This highlights the challenge of maintaining blinding in digital health interventions that involve active engagement. This is mitigated by the inclusion of the random nudge group, which acts as a useful unblinded comparator.

The reliance on Fitbit data for step count, while pragmatic for a large-scale study, means our findings are specific to this measurement method. The 7-day/500-step compliance criterion served as a proxy for wear-time, but direct wear-time metrics could provide more granular insights in future studies.

# Limitations and Future Directions

The primary endpoint's lack of statistical significance at two months, despite positive trends at one month and in longitudinal models, suggests that effects may not be expected to last long. Future directions should explore the strategies to improve and sustain engagement and efficacy over longer durations.

The nominal p-values for secondary outcomes should be interpreted with caution. Future research should consider pre-registering a comprehensive plan for handling multiple comparisons for secondary outcomes.

Further research is needed to explore the specific COM-B components that are most amenable to dynamic nudging and how their effectiveness changes over time or across different user contexts. While our RL algorithm adapted content and timing, future iterations could explore expansions, such as: other intervention parameters (e.g., message frequency, intensity, modality), dynamic components within the nudge text itself (e.g., reference prior step counts, trends, or patterns), or more diverse data streams (e.g., weather, social context, mood) into the RL state space to further enhance personalization. Future iterations should also explore the use of AI to develop nudges that are uniquely tailored to the individual.

# Conclusion

The PEARL study provides compelling evidence for the feasibility and early efficacy of a large-scale, behavioral science-informed reinforcement learning approach in digital physical activity interventions. While the two-month primary outcome was not statistically significant, the significant short-term gains and sustained effects observed in the RL arm highlight the power of adaptive algorithms to design effective digital interventions by personalizing and optimizing health behavior change. This work contributes significantly to the growing body of literature on JITAIs and underscores the importance of dynamic, data-driven personalization for scalable and effective digital health solutions.

# Methods

## Study design

Trial design details have been previously publicly registered and are available at https://osf.io/tw7up. The protocol was designed, written, and executed by the investigators. Study enrollment began in February 2024 and completed in May 2024. Follow-up of all patients ended in July 2024; the final study database was available in August 2024.

## Interventions

Interventions within the PEARL study comprised a combination of a nudge theme and a delivery time, resulting in a total of 12 distinct action classes in each intervention group. The six nudge themes included Ability, Perceived Benefit, Physical Opportunity, Planning, Prioritization, and Social Opportunity. Delivery times were set for morning (6 AM) and afternoon (3 PM). Upon selection of an action class, a specific message was randomly sampled from the corresponding theme's repository for display to the user, to ensure diversity and minimize habituation patterns. For each theme, approximately 20 messages were initially developed by behavioral science experts, supplemented by an additional 10 messages generated using large language models (LLMs) and subsequently validated by behavioral experts. Participants in the intervention arms (excluding the control group) received one text-only in-app push notification daily, drawn from this message repository, which contained approximately 180 total messages. Nudges were designed to be dismissible and were presented once per day. However, all delivered nudges remained accessible on the study home page within the Fitbit application throughout the day. Furthermore, participants were provided with the option to offer binary feedback (thumbs up/down) and submit free-text comments on each nudge.

## Fixed arm logic

The fixed arm message selection logic was designed to personalize nudge content and timing based on initial participant surveys. Upon study onboarding, participants completed a demographics survey, a COM-B survey, and a nudge time preference survey. For nudge content selection, mean scores for the six COM-B sub-themes were computed from the user's Likert scale ratings (1-5), with higher scores indicating that these themes represented *less* of a barrier to physical activity. A "barrier score" for each theme was then calculated as 5 minus the Likert score. For each day of the 60-day study, a nudge theme was sampled from a multinomial distribution weighted by these barrier scores, meaning themes associated with higher perceived barriers were more likely to be selected. For nudge time selection, participants' stated preferences (morning, afternoon, or no preference) determined the probability of a morning or afternoon nudge (e.g., 70% for preferred time, 30% for the other, or 50% for no preference). A time was then sampled daily from a binomial distribution based on these probabilities. This static policy, derived solely from initial survey responses, remained fixed for each participant throughout the entire study duration, with no further adaptation based on real-time user behavior or feedback.

## Reinforcement learning policy

The adaptive RL arm was designed according to a contextual multi-armed bandit (C-MAB) framework[19]. This specific subclass of RL well suits mHealth settings[20] and has been employed in previous studies[15,21]. C-MAB strategies are ideal candidates for capturing rapid changes in an individual user's context and needs, with the scope of efficiently delivering JITAIs in dynamic environments. Furthermore, as opposed to other RL classes, C-MABs enjoy a reduced computational burden from carrying through only the most recent information. We refer to Deliu at al. (2024) for the technical details[19].

At a pilot phase, various C-MAB algorithms have been explored, suggesting e-greedy as the optimal compromise between ease of interpretation, computational feasibility and a simple yet efficient balance between exploitation and exploration[11]. E-greedy trades-off exploitation and exploration as follows: with probability $1-\epsilon$ the policy's recommendation (i.e., the action that maximizes the expected reward) is followed, and with probability $\epsilon$ an action is chosen uniformly at random from the set of 12 possible actions. Clearly, a lower $\epsilon$ leads to higher exploration. In this work, we set $\epsilon$ as one of 0.7 or 0.8 throughout the study; the higher exploration value is chosen when the model deployed has high variability in decision across ensemble models.

There three main components that characterize the RL framework are the state, action and reward variables. In the context of PEARL, the state encapsulates user-specific information, while the action corresponds to the specific combination of the intervention component and delivery time of a nudge. Finally, the reward is the *proximal outcome* defined upon the step counts and serves as a proxy for the user's physical activity level. Below, we provide a formalization of these three main ingredients, contextualizing them within the PEARL study, followed by a description of the RL deployment pipeline.

**State variables**. These characterize the user's state on each day *t* and define the feature set used for *personalization* purposes. It comprises static information (demographics, surveys, baseline steps) as well as the individual time-varying history (past walking trends, app activity, past nudge feedbacks, etc.). We use $\mathbf{x}_{i,t}$ to denote the features of user $i$ on their day $t$ of study. The full set of information is given in Table 7 below.

**Table 7. Feature set for PEARL Reinforcement Learning Algorithm**

| User Feature class | State class | Cadence |
|---|---|---|
| User Demographics | Age, Gender, Device Type | Static: Start of study |
| | Education, Area type, weather | Start of study |
| User COM-B Survey | 20 survey questions with 1-5 likert scale | Start of study |
| User Pre Study steps | Pre-study steps mean and standard deviation | Start of study |
| | Pre-study walk pattern : Low vs High walker, Regular vs Irregular walker across days of week | Start of study |
| User Dynamic features | Recent steps mean and standard deviation | Last 7 days |
| | Slope of regression line fitting recent steps | Last 7 days |
| | Daily nudge feedback stats | Last 7 days |
| | Day of the week | Each day |
| | Recent Study Walk Pattern : : Low vs High walker, Regular vs Irregular walker across days of week | Last 7 days |
| | Avg steps in Morning(<12pm) Evening (>12pm) | Last 7 days |
| | Percentage of missing data in steps | Last 7 days |
| | Past Week Nudge History | Each day |

**Action or intervention space**. This represents the intervention set and is defined upon 6 nudge themes (Ability, Perceived Benefit, Physical Opportunity, Planning, Prioritization, Social Opportunity) and 2 delivery times (morning, afternoon), for overall 12 actions. Let $a_{i,t}$ denote the specific action (among the 12 unique combinations) chosen by the RL policy user $i$ on their day $t$ of study. For more details, we refer to subsection *Interventions*. As opposed to the other study groups, the RL arm is designed to estimate and follow an action-selection policy that optimizes the cumulative reward over time across all users. Such a policy is described in subsection Policy learning.

**Reward variable.** This represents the *proximal outcome* used to guide the *optimization* of the intervention delivery. It is defined as the *relative change* in the number of steps walked between nudge delivery and the following 24 hours, relative to an individual pre-study baseline walking pattern. Formally, denoted by $\mathbb{I}(\text{morning}_{i,t})$ and $\mathbb{I}(\text{weekday}_t)$ the binary variables indicating whether the nudge shown to user $i$ on their day $t$ of study is a morning or an evening nudge, and whether day $t$ is a weekday or a weekend, respectively, the baseline walking pattern, say $y_{i,0}\left(\mathbb{I}(\text{morning}_{i,t}), \mathbb{I}(\text{weekday}_t)\right)$. This baseline represents the average daily step count within a 24-hour period starting from morning for user $i$ and is calculated over a 30-day window prior to the start of the study for each combination of $\mathbb{I}(\text{morning}_{i,t})$ and $\mathbb{I}(\text{weekday}_t)$. Now, denoted by $y_{i,t}$ the steps taken by user $i$ on day $t$ of the study within a 24-hour window starting from the time of the nudge on day $t$, the reward is defined as follows:

$$r_{i,t}(a_{i,t}) = \frac{y_{i,t} - y_{i,0}\left(\mathbb{I}(\text{morning}_{i,t}), \mathbb{I}(\text{weekday}_t)\right)}{y_{i,0}\left(\mathbb{I}(\text{morning}_{i,t}), \mathbb{I}(\text{weekday}_t)\right)}.$$

According to these three key components, the RL decision-making process is summarized in Fig 12 below. The diagram depicts the RL deployment pipeline using the data from the RL arm to estimate the best action (nudge theme and time combination) for user $i$ on day $t$ of the study. This pipeline is run daily at midnight, with the model decision for each user defined by 5am.

**Figure 13. PEARL Reinforcement Learning Deployment Pipeline**

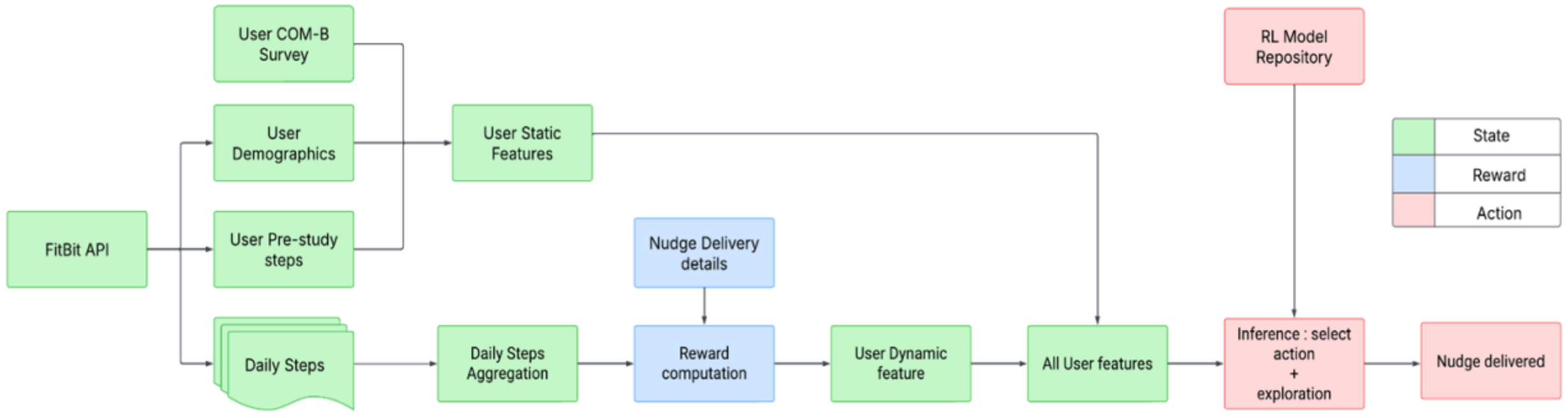

| | | | | | | | |
|---|---|---|---|---|---|---|---|
| | | | | | | | |
| | | | | | | | |

RL decision making pipeline

## Policy Learning

A decision-making or action-selection policy maps individual user states or contexts to an action. In RL arm, baseline and time-varying users' data are used to learn nudge policies and guide the action selection so as to optimize the expected reward. That is, given $\mathbf{x}_{i,t}$, the state of a user $i$ state on day $t$, we learn a mapping $\phi$ from the state space to the action space. The mapping is often characterized by a set of unknown parameters $W$, i.e., $\phi = \phi(\cdot; W)$. Denoting by $\phi(a; \mathbf{x}_{i,t}, W)$ the probability of playing action $a$ for a user $i$ with context $\mathbf{x}_{i,t}$, learning an optimal policy in this case is equivalent to estimating the unknown parameters such that:

$$W^* = \arg\max_W \sum_{i=1}^{N} \sum_{t=1}^{T} \mathbb{E}_{a \sim \phi(\cdot; x_{i,t}, W)}[r_{i,t}(a)]$$

In order to solve the optimization program, the expectation of $r_{i,t}(a_{i,t})$ was estimated using the history of previous data (states observed, nudges delivered and corresponding rewards) and importance sampling, a classical unbiased estimator of the value of a policy.
The optimization problem was tackled with an empirical risk minimization oracle which is readily available for various model classes such as decision trees, neural networks. After estimating the reward model for each of the 12 actions, an optimal action for user $i$ on day $t$ is chosen as $\hat{a}_{i,t} = \text{argmax}_a \mathbb{E}[\hat{r}_{i,t}(a)]$. XGBoost decision tree is used to implement reward prediction models.

## Statistical analysis

The overall demographics of the full sample and by each treatment group (no nudge, random nudge, fixed nudge, and RL nudge) were reported. Overall step counts within the sample were also presented descriptively. Balance of baseline characteristics was examined across the four study groups Difference-in-differences (DiD) regression analysis was conducted to determine the impact of the treatment arms on the change in step count for each participant. Benjamini-Hochberg adjustment was applied to control for Type I error across these pre-specified primary comparisons. Nominal p-values are reported for the secondary outcomes.

A generalized estimating equation (GEE) model with an exchangeable correlation structure was used to estimate the average effect of each treatment group on the daily step count over time, accounting for the longitudinal nature of the data and within-subject correlation.

All data analyses, including descriptive statistics, difference-in-differences regression analyses, and generalized estimating equation models, were performed using R (version 4.0.1, http://www.r-project.org).

# Data availability

The datasets generated and/or analyzed during the current study are not publicly available due to proprietary restrictions by the collaborating private company. Access to the data is governed by agreements with the company and cannot be shared without their explicit permission. Code and data were reviewed by Nina Deliu and Sofia Villar (both employed by Google at the time of review), who also verified the reproducibility of the main results.

# Code availability

The underlying code for this study [and training/validation datasets] is not publicly available for proprietary reasons. Code and data were reviewed by Nina Deliu and Sofia Villar (both employed by Google at the time of review), who also verified the reproducibility of the main results.